\documentclass{article}
\PassOptionsToPackage{numbers}{natbib}
\usepackage[final]{nips_2017}

\usepackage{fancyvrb}
\usepackage{xcolor}

\usepackage[utf8]{inputenc} 
\usepackage[T1]{fontenc}    
\usepackage{hyperref}       
\usepackage{url}            
\usepackage{booktabs}       
\usepackage{amsfonts}       
\usepackage{nicefrac}       
\usepackage{microtype}      
\usepackage{graphicx}
\usepackage{subcaption}

\title{Rasa: Open Source Language Understanding and Dialogue Management}

\author{
  Tom Bocklisch \\
  Rasa \\
  \texttt{tom@rasa.ai} \\
  \And
  Joey Faulkner \\
  Rasa \\
  \texttt{joey@rasa.ai} \\
  \And
  Nick Pawlowski \\
  Rasa \\
  \texttt{nick@rasa.ai} \\
  \And
  Alan Nichol\\
  Rasa\\
  \texttt{alan@rasa.ai} \\
}

\begin{document}

\maketitle

\begin{abstract}
  We introduce a pair of tools, Rasa NLU and Rasa Core, which are open
  source python libraries for building conversational software.
  Their purpose is to make machine-learning based dialogue management and language understanding
  accessible to non-specialist software developers.
  In terms of design philosophy, we aim for ease of use, and bootstrapping from minimal (or no)
  initial training data. 
  Both packages are extensively documented and ship with a comprehensive suite of tests. The code is available at \url{https://github.com/RasaHQ/}
\end{abstract}

\section{Introduction}

Conversational systems are becoming pervasive as a basis for human computer interaction as we seek more natural ways to integrate automation into everyday life. Well-known examples of conversational AI include Apple's Siri, Amazon's Alexa and Microsoft's Cortana but conversational systems are becoming widespread with platforms like Facebook Messenger opening up to chatbot developers. Common tasks for conversational systems include scheduling meetings\footnote{e.g. Meekan, see \url{https://meekan.com}}, booking flights\footnote{e.g. KLMs BlueBot, see \url{https://bb.klm.com/en}} and customer support tasks\footnote{see \url{http://rasa.ai/enterprise/case-studies/}}.

Modern open source libraries are held to a high standard of professionalism, and this extends to implementations of machine learning algorithms. There is a large amount of non-research work involved in maintaining a widely used project, and code produced by research groups often falls short of expectations\footnote{see \url{https://www.reddit.com/r/MachineLearning/comments/6l2esd/d_why_cant_you_guys_comment_your_fucking_code/}}. Rasa NLU and Core aim to bridge the gap between research and application, bringing recent advances in machine learning to non-experts who want to implement conversational AI systems.

We introduce Rasa NLU and Core as easy to use tools for building conversational systems, since until now there was no widely-used statistical dialogue system intended for non-specialists. Rasa is already used by thousands of developers worldwide. As with many other conversational systems, our tools are split into natural language understanding (Rasa NLU) and dialogue management (Rasa Core). Section \ref{sec:code_description} describes the code architecture, in \ref{sec:usage} we outline the developer experience, and in \ref{sec:demonstration} demonstrate an example application.
\newpage
\section{Related Work}
\label{sec:related_work}

Rasa takes inspiration from a number of sources.

Rasa's API uses ideas from scikit-learn \citep{pedregosa2011scikit} (focus on consistent APIs over strict inheritance) and Keras \citep{chollet2015keras} (consistent APIs with different backend implementations), and indeed both of these libraries are (optional) components of a Rasa application.

Text classification is loosely based on the fastText approach \citep{joulin2016bag}. Sentences are represented by pooling word vectors for each constituent token.
Using pre-trained word embeddings such as GloVe \citep{pennington2014glove}, the trained intent classifiers 
are remarkably robust to variations in phrasing when trained with just a few examples for each intent. 
Braun et al \citep{braun2017evaluating} show that Rasa NLU's performance compares favourably to various closed-source solutions.
Custom entities are recognised using a conditional random field \citep{lafferty2001conditional}.

Rasa Core's approach to dialogue management is most similar to \citep{williams2017hybrid},
but takes a different direction from many recent research systems.
There is currently no support for end-to-end learning, as in \citep{wen2016network} or \citep{DBLP:journals/corr/BordesW16}, where natural language understanding, state tracking, dialogue management, and response generation are all jointly learned from dialogue transcripts.
Rasa's language understanding and dialogue management are fully decoupled.
This allows Rasa NLU and Core to be used independently of one another, and allows trained dialogue models to be reused across languages.
For language generation, we encourage developers to generate a variety of responses by authoring multiple templates for each response. 
This is currently easier and more reliable than using (for example) a neural network to generate grammatically coherent and semantically correct responses as in \citep{wen2015semantically}.
Rasa Core also does not account for uncertainty in voice transcription and NLU, as would typically be achieved with Partially Observable Markov Decision Processes (POMDP) \citep{young2013pomdp}.
While there is support for training via reinforcement learning (currently in alpha), we do not emphasise this for new users. 
Rather than implementing a reward function and a simulated user, or immediately placing humans in the loop, we encourage developers to train the dialogue policy interactively (see section \ref{sec:interactive_learning}).
Recent work by Williams follows a similar machine teaching approach \citep{williams2017demonstration}, one difference being that in that work the dialogue policy is directly exposed to the surface form of the user utterances, whereas in Rasa Core the dialogue policy only receives the recognised intent and entities.

\subsection{Related Software Libraries}
As mentioned in the introduction, the large body of high-quality research into statistical dialogue systems
from the last decades has not translated into widely used software libraries.
A notable contribution is to this end is PyDial \citep{ultes2017pydial}, a recently released toolkit for dialogue research. 
Compared to PyDial, Rasa Core emphasises the needs of non-specialist software developers over those of researchers in the field. 
Other open toolkits include OpenDial \citep{lison2016opendial} and RavenClaw \citep{Bohus2009Ravenclaw}.

There are a number of general-purpose natural language processing (NLP) libraries in widespread use, and it is not necessary
to mention them all here. 
There are also a number of online services for natural language understanding (NLU), as the term is understood in the dialogue research community: converting short user messages into dialogue acts comprising an intent and a set of entities. 
Like the online services, Rasa NLU hides implementation details from new users, but with the advantage that slightly more experienced users can fully customise their NLU system.
Language understanding is performed by a number of components implementing a common API, and is
therefore easily configurable to suit the needs of a particular project.

\section{Description of the Code}
\label{sec:code_description}
Rasa's architecture is modular by design. This allows easy integration with other systems. For example, Rasa Core can be used as a dialogue manager in conjunction with NLU services other than Rasa NLU. While the code is implemented in Python, both services can expose HTTP APIs so they can be used easily by projects using other programming languages.

\subsection{Architecture}
\label{sec:arch}

Dialogue state is saved in a tracker object. There is one tracker object per conversation session, and this is the only stateful
component in the system. 
A tracker stores slots, as well as a log of all the events that led to that state and have occurred within a conversation. 
The state of a conversation can be reconstructed by replaying all of the events.

When a user message is received Rasa takes a set of steps as described in figure 1. Step 1 is performed by Rasa NLU, all subsequent steps are handled by Rasa Core. 

\begin{figure}[h]
  \centering
  \includegraphics[width=\textwidth]{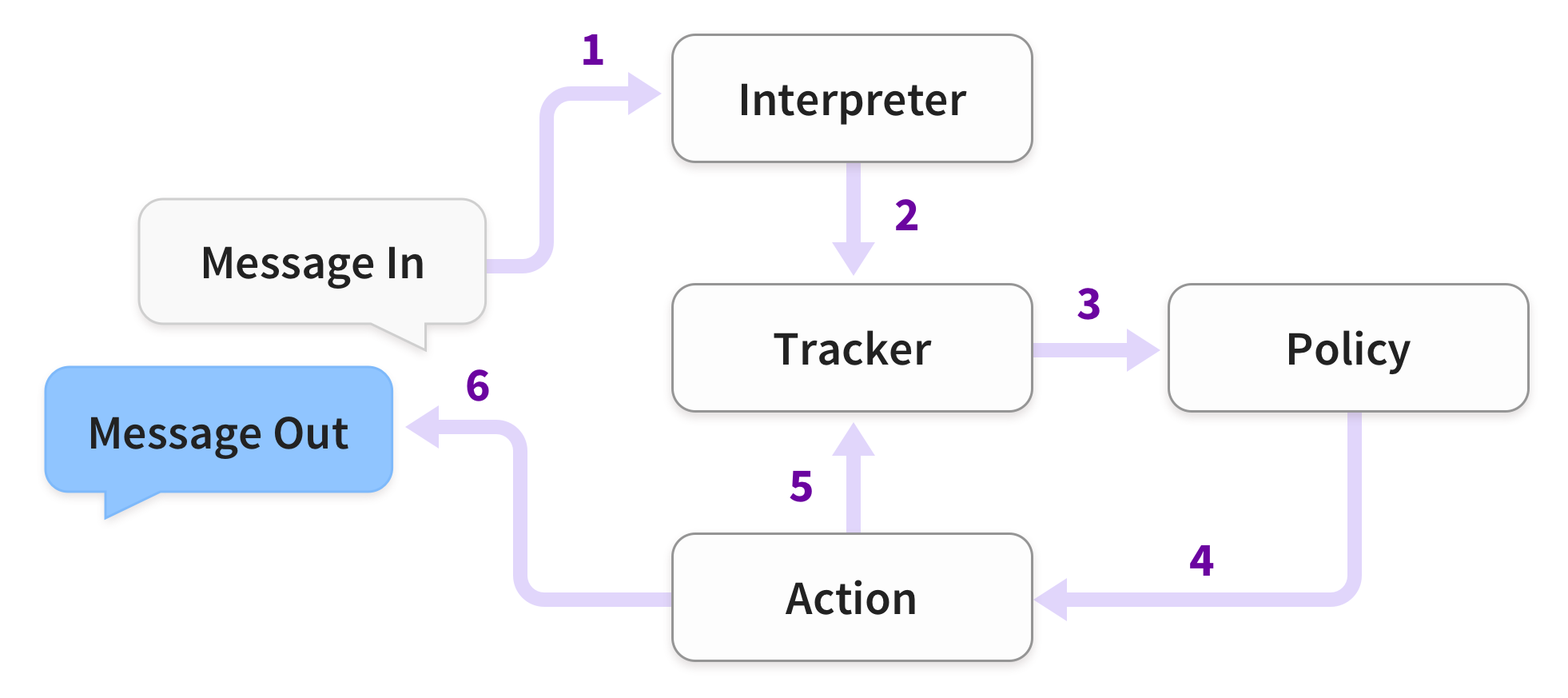}
  \label{message_flow}
  \caption{ 1. A message is received and passed to an Interpreter (e.g. Rasa NLU) to extract the intent, entities, and any other structured information. 2. The Tracker maintains conversation state. It receives a notification that a new message has been received. 3. The policy receives the current state of the tracker. 4. The policy chooses which action to take next. 5. The chosen action is logged by the tracker. 6. The action is executed (this may include sending a message to the user). 7. If the predicted action is not `listen', go back to step 3. }
\end{figure}

\subsection{Actions}
\label{sec:action_definitions}

We frame the problem of dialogue management as a classification problem.
At each iteration, Rasa Core predicts which action to take from a predefined list.
An action can be a simple utterance, i.e. sending a message to the user, or it can
be an arbitrary function to execute.
When an action is executed, it is passed a tracker instance, and so can make use of any
relevant information collected over the history of the dialogue: slots, previous utterances,
and the results of previous actions. 

Actions cannot directly mutate the tracker, but when executed may return a list of events.
The tracker consumes these events to update its state. 
There are a number of different event types, such as 
\verb|SlotSet|, \verb|AllSlotsReset|, \verb|Restarted|, etc.
The full list is in the documentation at \url{https://core.rasa.ai}.

\subsection{Natural Language Understanding}
\label{sec:nlu_pipelines}

Rasa NLU is the natural language understanding module.
It comprises loosely coupled modules combining a number of natural language processing and machine learning
libraries in a consistent API.
We aim for a balance between customisability and ease of use. 
To this end, there are pre-defined pipelines with sensible defaults which work well for most use cases.
For example, the recommended pipeline, \verb|spacy_sklearn|, processes text with the following components. First, the text is tokenised and parts of speech (POS) annotated using the spaCy NLP library. Then the spaCy featuriser looks up a GloVe vector for each token and pools these to create a representation of the whole sentence. Then the scikit-learn classifier trains an estimator for the dataset, by default a mutliclass support vector classifier trained with five-fold cross-validation. The \verb|ner_crf| component then trains a conditional random field to recognise the entities in the training data, using the tokens and POS tags as base features. 
Since each of these components implements the same API, it is easy to swap (say) the GloVe vectors for custom, domain-specific word embeddings, or to use a different machine learning library to train the classifier. There are further components for handling out-of-vocabulary words and many customisation options for more advanced users. All of these are detailed in the documentation at \url{https://nlu.rasa.ai}.


\subsection{Policies}
\label{sec:policies}

The job of a policy is to select the next action to execute given the tracker object. 
A policy is instantiated along with a featurizer, which creates a vector representation of
the current dialogue state given the tracker. 

The standard featurizer concatentates features describing:

\begin{itemize}
\item what the last action was
\item the intent and entities in the most recent user message
\item which slots are currently defined
\end{itemize}

The featurization of a slot may vary.
In the simplest case, a slot is represented by a single binary vector element indicating whether it is filled.
Slots which are categorical variables are encoded as a one-of-k binary vector,
those which take on continuous values can specify thresholds which affect their featurisation, or simply be passed to the featurizer as a float.

There is a hyperparameter \verb|max_history| which specifies the number of previous states
to include in the featurisation. 
By default, the states are stacked to form a two-dimensional array, which can be processed by a 
recurrent neural network or similar sequence model.
In practice we find that for most problems a \verb|max_history| value between 3 and 6 works well.

\section{Usage}
\label{sec:usage}

\subsection{Training Data Formats}
\label{sec:data_formats}

Both Rasa NLU and Core work with human-readable training data formats. 
Rasa NLU requires a list of utterances annotated with intents and entities. 
These can be specified either as a json structure or in markdown format.
The markdown syntax is especially compact and easy to read and can be
rendered by many text editors and web applications like GitHub.

\begin{verbatim}
## intent: restaurant_search
- show me [chinese](cuisine) restaurants
\end{verbatim}

The json format is slightly more cumbersome to read, but is not whitespace sensitive and more suitable for transmission of training data between applications and servers.  
\begin{verbatim}
{
  "text": "show me chinese restaurants",
  "intent": "restaurant_search",
  "entities": [
    {
      "start": 8,
      "end": 15,
      "value": "chinese",
      "entity": "cuisine"
    }
  ]
}
\end{verbatim}

Rasa Core employs markdown to specify training dialogues (aka `stories'). 

\begin{verbatim}
## story_07715946
* greet
   - utter_ask_howcanhelp
* inform{"location":"rome","price":"cheap"}
   - utter_on_it
   - utter_ask_cuisine
* inform{"cuisine":"spanish"}
   - utter_ask_numpeople
* inform{"people":"six"}
   - action_ack_dosearch
\end{verbatim}

A story starts with a name preceeded by two hashes \begin{verbatim}## story_03248462\end{verbatim}
The choice of name is arbitrary but can be helpful for debugging.
The body of a story is a sequence of events, separated by newlines. 
An event such as \begin{verbatim}* inform{"location":"bombay", "price":"expensive"}\end{verbatim} 
is a user utterance, annotated as a dialogue act. 
The general format is \begin{verbatim}* intent{"entity_name": "entity_value"}\end{verbatim} 
where entities are key-value pairs separated by commas.
System actions are also events and are specified by lines starting with a dash. 
The end of a story is denoted by a newline.

\subsection{Machine Teaching}
\label{sec:interactive_learning}

In addition to supervised learning, Rasa Core supports a machine teaching approach where 
developers correct actions made by the system.
We find that this is a practical approach for generating training data, and exploring the space
of plausible conversations efficiently. 

Here is an example of a user engaged in machine teaching of a restaurant recommender system (described in Section \ref{sec:demonstration}) . 
The user is presented with the following prompt\footnote{In Rasa a conversational system is referred to as a `bot'}:
\begin{samepage}
\begin{Verbatim}[commandchars=\\\{\}]
------
Chat history:

    bot did:	[]

    bot did:	action_listen

    user said:	\textcolor{green}{/greet{}}

        whose intent is:	greet

we currently have slots:  cuisine: \textcolor{blue}{None}, people: \textcolor{blue}{None},
            price: \textcolor{blue}{None}, location: \textcolor{blue}{None}

------
The bot wants to [\textcolor{red}{utter_ask_howcanhelp}] due to the intent. Is this correct?

	1.	Yes
	2.	No, intent is right but the action is wrong
	3.	The intent is wrong
	0.	Export current conversations as stories and quit
>>
\end{Verbatim} 
\end{samepage}

If the user inputs \verb|2|, indicating that the action is wrong, they will be provided with a prompt as below which lists the possible actions and the probability assigned to them by the dialogue policy:
\begin{samepage}
\begin{verbatim}
------
what is the next action for the bot?

         0                           action_listen    0.12                          
                 ...
         8                       utter_ask_cuisine    0.03
         9                      utter_ask_helpmore    0.03
        10                    utter_ask_howcanhelp    0.19
        11                      utter_ask_location    0.04
        12                   utter_ask_moreupdates    0.03
        13                     utter_ask_numpeople    0.05
        14               action_search_restaurants    0.03
                ...
>>
\end{verbatim}
\end{samepage}
Choosing the correct action creates a new training data point. Rasa Core then partially trains the dialogue policy and moves the conversation forward to the next step. Once completed, the trained model is persisted and the newly generated training data is saved to a file. 
\subsection{Visualisation of Dialogue Graphs}
\label{sec:graphs}

\begin{figure}[ht]
  \centering
  \begin{subfigure}{.5\textwidth}
    \centering
  	\includegraphics[width=.98\textwidth]{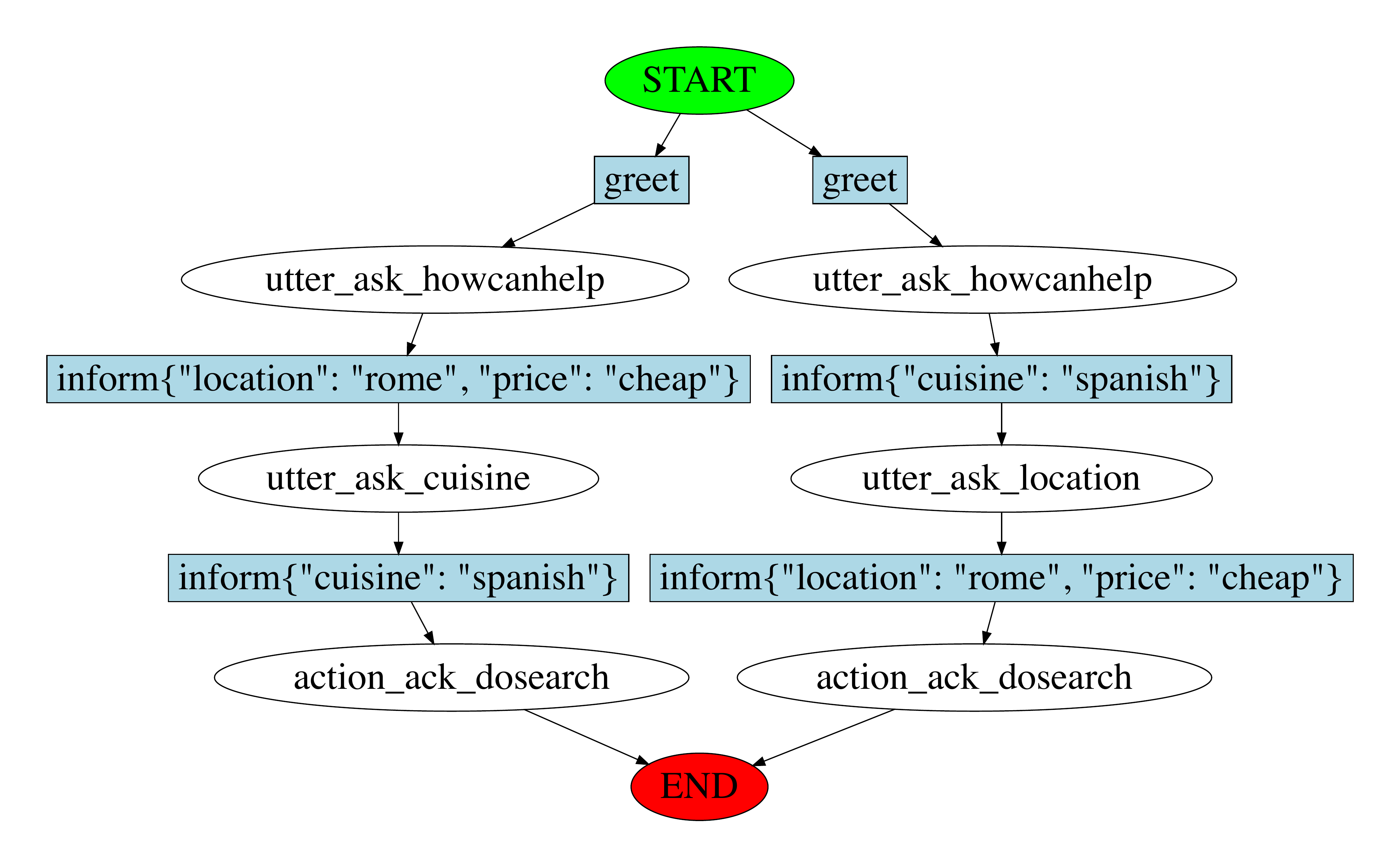} 
  	\caption{Example story graph without simplification}
  \end{subfigure}%
  \begin{subfigure}{.5\textwidth}
  	\centering
  	\includegraphics[width=.98\textwidth]{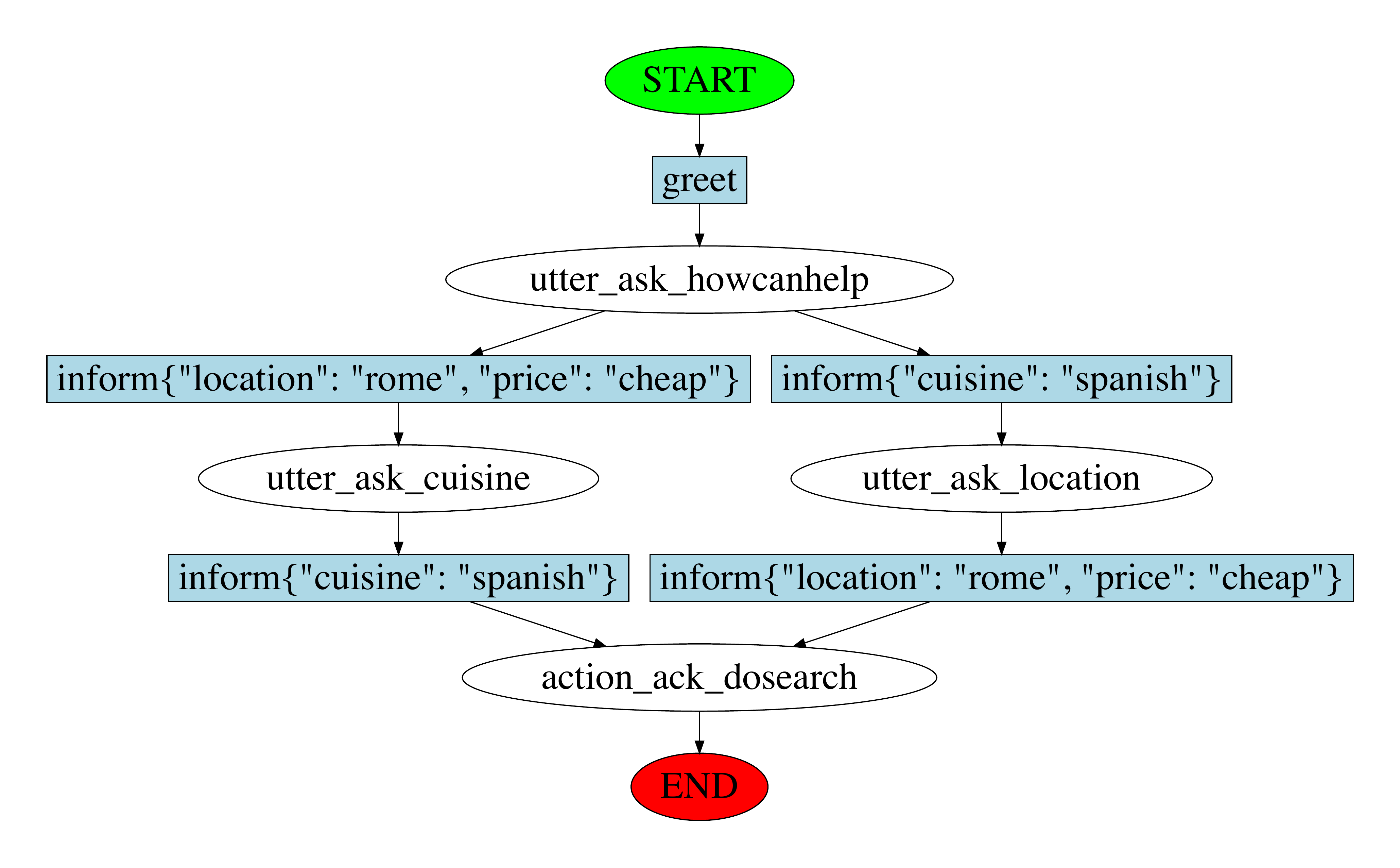}
  	\caption{Example story graph with simplification}
  \end{subfigure}
  \caption{Minimal example illustrating how story graphs are simplified. The training data contains two stories with the same first interaction. These nodes are therefore considered equivalent and merged.}
  \label{fig:story_graph}
\end{figure}

Rasa Core also has the capability to visualise a graph of training dialogues. 
A story graph is a directed graph with actions as nodes. Edges are labeled with the user utterances that occur in between the execution of two actions. If there is no user interaction between two consecutive actions, the edge label is omitted.  
Each graph has an initial node called \verb|START| and a terminal node called \verb|END|.
Note that the graph does not capture the full dialogue state and not all possible walks along the edges necessarily occur in the training set.
To simplify the visualization a heuristic is used to merge similar nodes. A generated graph before and after running the simplification is shown in Figure \ref{fig:story_graph}. During the simplification two nodes are merged by replacing them with a single node that inherits all the incoming and outgoing edges, removing duplicates in the process. This makes the resulting graphs easier to interpret.

Nodes are merged if these two conditions are met:
\begin{itemize}
  \item they represent the same action
  \item the previous \verb|max_history| turns are identical for all dialogues leading to the nodes.
\end{itemize}

\subsection{Deployment in a Production Environment}

The repositories for Rasa NLU and Core both contain Dockerfiles for producing static virtual machine (VM) images. This aids in reproducibility and ease of deployment to a variety of server environments. The web servers running the HTTP API support both thread-based and process-based parallelism, allowing them to handle large request volumes in a production environment.

\section{Demonstration}
\label{sec:demonstration}
To demonstrate the usage of Rasa Core, we use the BAbl dialogue dataset \citep{DBLP:journals/corr/BordesW16}. This is a simple slot-filling exercise where the system is asked to search for a restaurant and has to fill several slots before being able to perform a successful search. The system may ask the user for their preference for any slot. The available slots are location, the number of people, the cuisine and the price range.

This is an interesting dataset due to the inherent non-linearity of the problem - there are multiple ways to get the same information so there isn't a single `correct'  action in every case. Accuracy and precision are therefore not the most appropriate metrics for evaluating a dialogue policy. Instead we consider how the system chooses actions depending on what information is already available. It should attempt to fill slots which are empty.

\begin{figure}
\includegraphics[width=\textwidth]{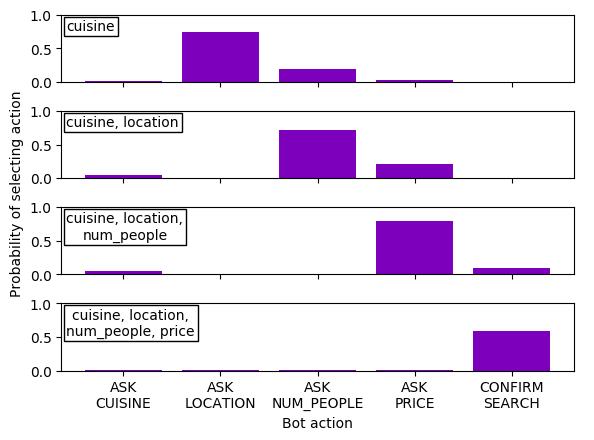}
\caption{Here we plot the probabilities of choosing actions for the bAbl example (Section~\ref{sec:demonstration}) . We sequentially inform the system of the correct (top-bottom): cuisine, location, number of people and price, and the system chooses the next action based on the information it already has (which is listed in the box on the top left). The system favours asking about slots it has not been informed of, until all of the slots are filled and it searches for the restaurant. } 
\label{slot_figure}
\end{figure}

In Figure~\ref{slot_figure} we can see the probabilities which Rasa Core attaches to each action given the slots it already knows. We see that Core follows a rough pattern of asking about cuisine, followed by location, followed by number of people - as it does in the training data. However it recognises that it could also ask about the other unfilled slots by attributing a non-zero probability to each one. The filled slots (the lower diagonal of the 	`grid') are given vanishingly small probability. This illustrates how Rasa Core can use contextual clues to learn non-linear conversations.

\section{Outlook}
\label{sec:outlook}

Both Rasa NLU and Core are under active development. 
They serve as a platform for making applied research in conversational AI usable by non-specialist developers, and as such will never be `finished'.
A number of topics are under active development, including improved support for reinforcement learning, making NLU robust to typos and slang, and supporting more languages. We also plan to release real-world datasets for comparing the performance of different models. The authors welcome external contributions to the project, the specifics of which can be found in the repositories on GitHub. 

\section*{Acknowledgements}
The authors are indebted to the users of both libraries for providing invaluable feedback and creating a supportive community around these tools. Special acknowledgement is owed to the external contributors to both libraries. Up-to-date lists of contributors may be viewed at \url{https://github.com/RasaHQ/rasa_nlu/graphs/contributors} and  \url{https://github.com/RasaHQ/rasa_core/graphs/contributors}.

\bibliographystyle{plain}
\bibliography{rasa_refs}

\end{document}